\documentclass[runningheads]{llncs}

\usepackage{graphicx}
\usepackage{comment}
\usepackage{amsmath,amssymb} 

\usepackage{cuted}
\usepackage[ruled, noend, noline]{algorithm2e}
\usepackage{tabularx}
\usepackage{subcaption}
\usepackage{lipsum}
\usepackage{multirow}
\usepackage{color,soul}
\usepackage{hyperref}
\hypersetup{linkcolor=red, colorlinks=true}
\usepackage[inline]{enumitem}
\usepackage{color}

\makeatletter
\@namedef{ver@everyshi.sty}{}
\makeatother

\def\ECCVSubNumber{3024}  

\titlerunning{Learning to Separate: Detecting Heavily-Occluded Objects in Urban Scenes}

\author{Chenhongyi Yang\inst{1[0000-0003-3895-6895]}\and
Vitaly Ablavsky\inst{2[0000-0003-2703-7666]}\thanks{Work performed at Boston University.} \and \\
Kaihong Wang\inst{1[0000-0002-0637-9862]} \and 
Qi Feng\inst{1[0000-0001-6342-3228]} \\ \and
Margrit Betke\inst{1[0000-0002-4491-6868]}}
\authorrunning{C. Yang \it{et al.}}

\institute{
Boston University\\
\email{\{hongyi,kaiwkh,fung,betke\}@bu.edu} \and 
University of Washington\\
\email{vxa@uw.edu}}

\title{Learning to Separate: Detecting Heavily-Occluded Objects in Urban Scenes}

\begin{document}

\pagestyle{headings}
\mainmatter

\newcommand{\convmap}{$F$}
\newcommand{\rpnbox}{$\mathbf{B}$}
\newcommand{\regoutput}{$R$}
\newcommand{\regbox}{$\Tilde{\mathbf{B}}$}
\newcommand{\aeoutput}{$AE$}
\newcommand{\clsoutput}{$C$}
\newcommand{\finalbox}{$\hat{\mathbf{B}}$}
\newcommand{\mmiou}{\textsc{MMIoU}}

\newcommand{\downplay}[1]{#1}


\newcommand{\agidsyb}{e}
\newcommand{\agidset}{\mathrm{E}}
\newcommand{\agsyb}{e}
\newcommand{\agidfull}{Semantics-Geometry Embedding (SGE)}
\newcommand{\agidshort}{Semantics-Geometry Embedding}
\newcommand{\agid}{SGE}
\newcommand{\agnmsfull}{Semantics-Geometry Non-Maximum-Suppression (SG-NMS)}
\newcommand{\agnmsshort}{Semantics-Geometry Non-Maximum-Suppression}
\newcommand{\agnms}{SG-NMS}
\newcommand{\agc}{SG-NMS-Constant}
\newcommand{\agl}{SG-NMS-Linear}
\newcommand{\ags}{SG-NMS-Square}

\maketitle

\begin{abstract}
  
  While visual object detection with deep learning has received much attention in the past decade, cases when heavy intra-class occlusions occur have not been studied thoroughly. In this work, we propose a Non-Maximum-Suppression (NMS) algorithm that dramatically improves the detection recall while maintaining high precision in scenes with heavy occlusions. Our NMS algorithm is derived from a novel embedding mechanism, in which the semantic and geometric features of the detected boxes are jointly exploited. The embedding makes it possible to determine whether two heavily-overlapping boxes belong to the same object in the physical world. Our approach
  is particularly useful for car detection and pedestrian detection in urban scenes where occlusions often happen.
  We show the effectiveness of our approach by creating a model called SG-Det (short for Semantics and Geometry Detection) and testing SG-Det on two widely-adopted datasets, KITTI and CityPersons for which it achieves state-of-the-art performance. Our code is available at \href{https://github.com/ChenhongyiYang/SG-NMS}{https://github.com/ChenhongyiYang/SG-NMS}.
\end{abstract}


\section{Introduction}
Recent years have witnessed significant progress in object detection using deep convolutional neural networks (CNNs)~\cite{deng2009imagenet,he2016deep,simonyan2014deep}. 
The approach taken by many state-of-the-art object detection methods~\cite{RedmonDiGiFa16,LiuAnErSzReFuBe16,GirshickDoDaMa14,girshick2015fast,RenHeGiSu15} is to predict multiple bounding boxes for an object and then use a heuristic method such as non-maximum suppression (NMS) to remove superfluous bounding boxes that stem from duplicate detected objects.

  The Greedy-NMS algorithm is easy to implement and tends to work well in images where objects of the same class do not significantly occlude each other. However, in urban scenes, where the task is to detect potentially heavily occluded cars or pedestrians, Greedy-NMS does not perform adequately.  The decrease in accuracy is due to the fundamental limitation of the NMS algorithm, which uses a fixed threshold to determine which bounding boxes to suppress: The algorithm cannot suppress duplicate bounding boxes belonging to the {\em same} object while preserving boxes belonging to {\em different} objects, where one object heavily occludes others. 
  Soft-NMS~\cite{BodlaSiChDa17} attempts to address this limitation by not removing overlapping boxes but instead lowering their confidence; however, all overlapping boxes are still treated as false positives regardless of how many physical objects are in the image.  

\begin{figure}[!t]
\centering
  \includegraphics[width=\linewidth]{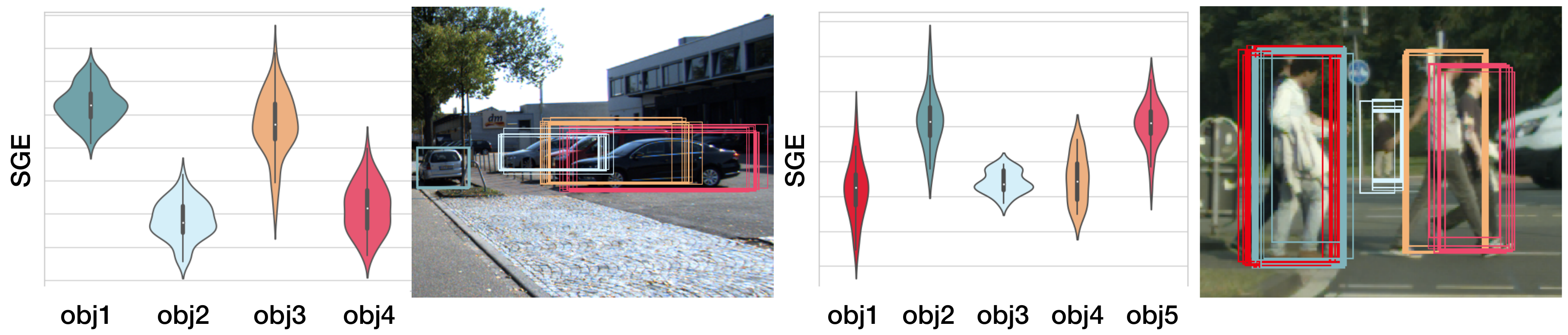}
  \caption{Learned \agidfull\ for bounding boxes predicted by our proposed detector on KITTI and CityPersons images. Heavily overlapped boxes are separated in the \agid \ space according to the objects they are assigned to. Thus, distance between \agid s \ can guide Non-Maximum-Suppression to keep correct boxes in heavy intra-class occlusion scenes. }
  \label{figure1}
\end{figure}

The limitation of NMS could be circumvented with an oracle that assigns each bounding box an identifier that corresponds to its physical-world object.
Then, a standard NMS algorithm could be applied per set of boxes with the same identifier (but not across identifiers), thus ensuring that false positives from one object do not result in suppression of a true positive from a nearby object. 

To approximate such an oracle, we can try to learn a mapping from boxes into a latent space so that the heavily overlapping boxes can be separated in that space. Naively, this mapping can be implemented by learning an embedding for every box based on its region features, e.g., the pooled features after RoIPooling~\cite{girshick2015fast}. However, the usefulness of such an embedding would be limited because heavily overlapping boxes tend to yield similar region features, thus would map to nearby points in the embedding space. In this paper, we demonstrate that by considering both the region features and the geometry of each box, we can successfully learn an embedding in a space where heavily overlapping boxes are separated if they belong to different objects. We call the learned embedding \agidfull. We also propose a novel NMS algorithm that takes advantage of the \agid \ to improve detection recall.

We visualize the concept of a  \agidfull \ in Fig.~\ref{figure1}, where boxes belonging to the same object are mapped to a similar \agid \ and boxes belonging to different but occluded objects are mapped to \agid s that are far away. Although the embedding algorithm may assign boxes in disparate parts of an image to similar \agid s, this does not negatively impact our \agnms \ algorithm because these boxes can be easily separated based on their intersection-over-union (IoU) score. The \agid \ is implemented as an associative embedding~\cite{newell2017associative} and learned using two loss functions, {\em separation} and {\em group loss}. To train the \agid \ with the object detector end-to-end, we propose a novel Serial Region-based Fully Convolutional Network (Serial R-FCN),  where the geometric feature of each detected box is precisely aligned with its semantic features. We combine this network with the \agnmsfull \ algorithm in a model we call SG-Det (short for Semantics and Geometry Detection).
In summary, we make three main contributions:
\begin{enumerate}[itemsep=-4pt,topsep=1pt,itemsep=0pt]
  \item
    A bounding-box-level {\em \agidfull} \ is proposed, and a Non-Maximum Suppression algorithm, called {\em \agnmsfull,} based on this embedding, is derived.  The algorithm markedly improves object detection in scenarios with heavy intra-class occlusions. 
  \item
   A {\em serial R-FCN} with self-attention in each head is presented that not only provides the
    ability to learn the above-mentioned \agid\ end-to-end, but also improves
    object detection accuracy. 
  \item
    The model {\em SG-Det} is proposed, which combines the serial R-FCN and the \agnms \ algorithm.  {\em SG-Det} achieves state-of-the-art performance on the tasks of car detection for the KITTI~\cite{geiger2012we} dataset and pedestrian detection for the  CityPersons~\cite{zhang2017citypersons} dataset 
    by dramatically improving the detection recall
    in heavily-occluded scenes.
    
\end{enumerate}

\section{Related Works}
\textbf{Object Detection.}
CNN-based object detectors can be divided into one-stage and two-stage approaches. One-stage detectors~\cite{RedmonDiGiFa16,LiuAnErSzReFuBe16,lin2017focal} directly predict the object class and the bounding box regressor  by sliding windows on the feature maps. Two-stage object detectors~\cite{girshick2015fast,RenHeGiSu15,DaiLiHeSu16,CaiVa18}, first compute regions of interest (RoIs)~\cite{RenHeGiSu15,uijlings2013selective,ZhangWeBiLeLi18,zitnick2014edge,he2015spatial} and then estimate the class label and bounding box coordinates for each RoI. Although the two-stage approaches often achieve higher accuracy, they suffer from low computational efficiency. R-FCN~\cite{DaiLiHeSu16} addresses this problem by replacing the computation in fully-connected layers with nearly cost-free pooling operations. 

\textbf{Non Maximum Suppression.}
NMS is widely used in modern object detectors to remove duplicate bounding boxes, but it may mistakenly remove boxes belonging to different objects. Soft-NMS~\cite{BodlaSiChDa17} was proposed to address this problem by replacing the fixed NMS threshold with a score-lowering mechanism. However, highly-overlapping boxes are still treated as false positives regardless of the semantic information. In Learning-NMS~\cite{HosangBeSc17}, a neural network is used to perform NMS, but the appearance information is still not considered. The Adaptive-NMS approach~\cite{LiuHuWa19} learns a threshold with the object detector, but when the threshold is set too high, false positives may be kept. The relation of bounding boxes can also be used to perform NMS by considering their appearance and geometric features~\cite{HuGuZhDaWe18}, but this does not handle intra-class occlusion. The localization quality of each box can be learned to help NMS with keeping accurate boxes~\cite{Tychsen_Smith_2018,HeZhWaSaZh19,jiang2018acquisition,Tan_2019_ICCV}.

\begin{figure*}[!t]
\begin{center}
    \includegraphics[width=\linewidth]{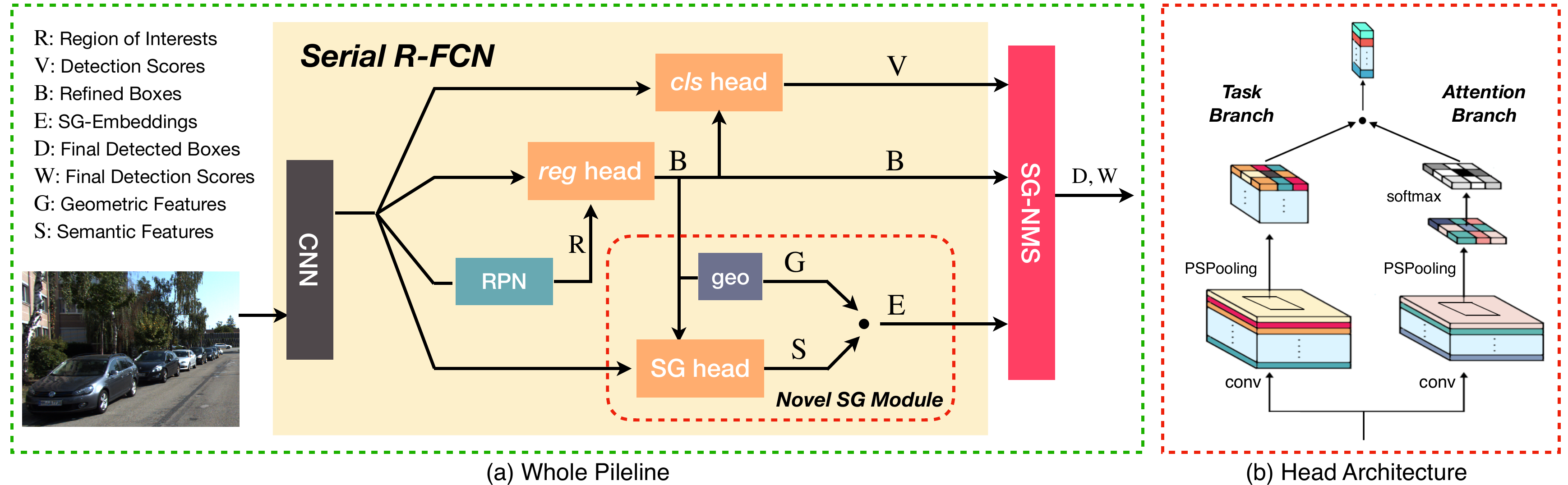}
\caption{(a) Overview of our proposed model SG-Det. An input image is first processed by a backbone CNN to yield feature maps.  A Region Proposal Network (RPN)~\cite{RenHeGiSu15} is used to extract regions of interests (ROIs). The RoIs will first be refined by the regression head and then fed into the classification head to produce detection scores, making the whole pipeline serial. A novel \textbf{Semantics-Geometry Module}, parallel to the classification head, is added to learn the SG embedding for each refined box. Finally, the detected box, detection scores, and SG embeddings are fed into the \agnms \ algorithm to produce final detections. 
(b) All heads (\textcolor[rgb]{1,0.68,0.44}{orange} boxes in Fig.~\ref{fig:pipeline}(a)) share a similar architecture. The feature map computed by the backbone network is processed by two branches to yield two score maps. Then a Position Sensitive RoI-Pooling~\cite{DaiLiHeSu16} is applied to produce two grids of $k^2$ position-sensitive scores -- a task-specific score and an attention score. A softmax operation transforms the attention score into a discrete distribution over the $k^2$ grids. Finally the $k^2$ task scores are aggregated by the attention distribution to yield the final output scores.}
\label{fig:pipeline}
\end{center}
\end{figure*}

\textbf{Other Occlusion Handling Approaches.}
There are many other methods designed to handle occlusion, including both intra-class or inter-class occlusion. Most of them focus on detecting pedestrians in crowd scenes. Repulsion loss~\cite{WangXiJiShSuSh18} was proposed to prevent boxes from shifting to adjacent objects. The occluded person is detected by considering different body parts separately~\cite{ZhangWeBiLeLi18,zhang2018occluded,NohLeKiKi18,tian2015deep,zhou2017multi}. 
A novel expectation-maximization merging unit was proposed to resolve overlap ambiguities
~\cite{Goldman_2019_CVPR}.
Additional annotations such as head position or visible regions have been used \cite{ZhouYu18,pang2019mask,zhang2019double} to create robust person detectors. Although these approaches have been shown to be effective in detecting occluded persons, it is difficult to generalize them to other tasks like car detection.

\section{Methodology}

In this section, we first introduce the proposed Semantics-Geometry Embedding (Section~\ref{sec:eag1}), then the Semantics-Geometry NMS algorithm (Section~\ref{sec:eag2}), and finally the proposed Serial R-FCN (Section~\ref{sec:eag3}). The overview of the combined proposed model {\em SG-Det} is shown in Figure~\ref{fig:pipeline}.\\
See also \href{http://www.cs.bu.edu/fac/betke/research/SG-Det}{http://www.cs.bu.edu/fac/betke/research/SG-Det}.

\subsection{Semantics-Geometry Embedding}\label{sec:eag1}
Our key idea for separating occluded objects in an image is to map each putative detection to a point in a latent space. In this latent space, detections belonging to the same physical object form a tight cluster; detections that are nearby in the image plane, but belong to different physical objects, are pushed far apart.

To implement this idea, we design an embedding for each bounding box that takes the form of a dot-product
\begin{align}
    \agidsyb=\mathbf{s}^{T} \cdot \mathbf{g},
    {\label{eq:eag}}
\end{align}
where $\mathbf{g}$ is the {\em geometric feature} and $\mathbf{s}$ is the {\em semantic feature}. The geometric feature has a fixed form $\mathbf{g}=(x,y,w,h)^T$ 
with center coordinates $(x,y)$ and 
width and height $(w,h)$ of the bounding box. We tried different kinds of geometric features (e.g., \cite{HuGuZhDaWe18}) and feature vectors with higher dimensions produced by a fully-connected layer, but found that such complexity did not provide further significant improvement.

Unlike the geometric feature, the semantic feature $\mathbf{s}$ is a weight output by a function that yields a vector compatible with $\mathbf{g}$; the function is implemented as a neural network, as shown in the Semantics-Geometry Head in Fig.~\ref{fig:pipeline}. Note that the \agid\ is computed by the linear transformation of the geometric feature taking the learned semantic feature as a weight. An interpretation is that the neural network automatically learns how to distinguish the bounding boxes belonging to different objects. Note that a similar idea was proposed that combined geometric and semantic features in a Relation Network~\cite{HuGuZhDaWe18}, but our approach is much simpler and can handle intra-class occlusion effectively.

We train the \agid \ using the loss function defined in Eq.~\ref{eq:eag}. The training is carried out end-to-end, jointly with the object-detection branch using the loss function defined later (Eq.~\ref{eq:total-loss}).

The loss function is derived for the \agid \ by extending the notion of an {\em associative embedding}~\cite{newell2017associative,law2018cornernet}. Specifically, we use a {\em group loss} to group the \agid s of boxes belonging to the same object, and use a {\em separation loss} to distinguish \agid s of boxes belonging to different objects. For one image, the ground-truth boxes are denoted by $\mathrm{B}^*=\{b_1^*,b_2^*,...,b_M^*\}$. For each refined box $b_i$ in the refined box set  $\mathrm{B}=\{b_i\}$, let $b_j^*$ be the ground truth box with the largest IoU.
If $\textsc{IoU}(b_i,b_j^*)>\theta$, box $b_i$ would be ``assigned'' to $b_j^*$. Thus the refined bounding boxes are divided into $M+1$ sets:
$\mathrm{B}=\mathrm{B}_1\cup \mathrm{B}_2\ \cup ,..., \cup\
\mathrm{B}_{M+1}$, where $\mathrm{B}_{M+1}$ is the set of refined boxes that
are not assigned to any ground truth box. Then the {\em group} and {\em separation losses} are defined as:
\begin{align}
  L_\text{group}(\{\agidsyb_i\})&=\sum_{j=1}^{M}\sum_{b_i\in
  \mathrm{B}_j}{|\agidsyb_i-{\agidsyb_j}^*|},\\
  L_\text{sep}(\{\agidsyb_i\})&=\sum_{i}{p_i^* \max(0,
  \sigma-|\agidsyb_i-\Tilde{\agidsyb_i}|)},
\end{align} 
where ${\agidsyb_j}^*$ is the \agid\ of ground truth box $b_j^*$, $\Tilde{b_i^*}$ is the ground truth box with the second largest IoU with respect to $b_i$, and its \agid \ is $\Tilde{\agidsyb_i}$. We use $\sigma$  to stabilize the training process by preventing the distances between  embeddings to be infinite.  We found that the model performance is not sensitive to the actual value of $\sigma$. In the definition of the {\em separation loss}, $p_i^*$ is a indicator variable which is 1 only if $b_i\notin B_{M+1}$ and $\textsc{IoU}(b_i,\Tilde{b_i^*})>\rho$. 

\begin{algorithm}[!t]
\DontPrintSemicolon
  \SetKwInOut{Input}{Input}
  \Input{
  $\mathrm{B}=\{b_i\}$: List of detection locations (boxes),\\
  $\mathrm{V}=\{v_i\}$: List of detection scores,\\
  $\agidset=\{\agidsyb_i\}$: List of SGEs,\\
  $N_t$: \textsc{IoU} threshold,\\
  $\Phi(\cdot)$: $\mathbb{R} \rightarrow \mathbb{R}$, a monotonically increasing function}
  
  \Begin{
  $\mathrm{D}\xleftarrow[]{}\{\}$; 
  $\mathrm{W}\xleftarrow[]{}\{\}$  \tcp*{set D as detected boxes and W as their scores} 
  \While{$\mathrm{B}\neq  \emptyset$}{
  $m \xleftarrow{} \arg \max_{i\in \{1\ldots N\}} {\mathrm{V}}$\\
  $\mathrm{D}\xleftarrow{}\mathrm{D}\cup \{b_m\}$;
  $\mathrm{W}\xleftarrow{}\mathrm{W}\cup \{s_m\}$\\
  $\mathrm{B}\xleftarrow{}\mathrm{B}\setminus\{b_m\}$;
  $\mathrm{V}\xleftarrow{}\mathrm{V}\setminus\{v_m\}$\\
  \For{$b_i$ \textsc{in} $\mathrm{B}$}
  {
   $\tau \xleftarrow{} \textsc{IoU}(b_m,b_i)$\\
  \tcp{compare not only $\mathrm{IoU}$ but also the embedding distance}
  \If{$\tau \geq N_t$ \textsc{and} \textcolor{red}{$d(\agidsyb_m,\agidsyb_i)\leq \Phi(\tau)$}}
  {$\mathrm{B}\xleftarrow{}\mathrm{B}\setminus b_i$;
  $\mathrm{V}\xleftarrow{}\mathrm{V}\setminus v_i$}}}
  \Return $\mathrm{D}$, $\mathrm{W}$
  }
\caption{The proposed Semantics-Geometry NMS.}
\label{alg-ae-nms}
\end{algorithm}

Some readers may confuse our loss functions with the Repulsion Loss (RL)~\cite{WangXiJiShSuSh18}, which is completely different. The RL was proposed to improve bounding box regression so that the detected bounding boxes better fit ground-truth objects. In contrast, our method does not affect the bounding box regression. The embedding trained through the two loss functions is used to determine if two overlapping boxes belong to the same object. Another difference is that the RL is performed in the box-coordinate space, while our group and separation losses are performed in the latent embedding space.

\subsection{Semantics-Geometry Non-Maximum Suppression}\label{sec:eag2}
We now derive our simple, yet effective NMS algorithm, \agnms, \ which takes advantage of the \agidshort. Its pseudo code is given in Algorithm~\ref{alg-ae-nms}.

\agnms \ first selects the box with the highest detection score as the {\em pivot box}. For each of the remaining boxes, its IoU with the pivot box is denoted by~$\tau$, and the box will be kept if the $\tau < N_t$. When $\tau > N_t$, \agnms \ checks the distance between its \agid\ and the \agid of the pivot box. If the distance is larger than $\Phi(\tau)$, the box will also be kept. Here $\Phi(\cdot)$ is a monotonically increasing function, which means that, as $\tau$ increases, a larger distance is required to keep it. In this work, we consider three kinds of \agnms \ algorithms: \agc, \agl \ and \ags, which respectively correspond to:
\begin{align}
  \Phi(\tau) = t_c, \;\; \Phi(\tau) = t_l\cdot \tau , \;\;  {\mathrm{and}} \; \; \Phi(\tau) = t_s \cdot \tau^2,
\end{align}
where $t_c$, $t_l$, and $t_s$ are hyper-parameters.


\subsection{The proposed Serial R-FCN}\label{sec:eag3}
In order to compute \agid s that can capture the difference between geometric features of boxes belonging to different objects, we need to align extracted semantic features strictly with the refined boxes after bounding box regression. However, this cannot be achieved by normal two-stage CNN-based object detectors where the pooled feature is aligned with the RoI instead of the refined box because of the bounding-box regression. 

To address this problem, we propose {\em Serial} R-FCN, see Fig.~\ref{fig:pipeline} (a). In Serial R-FCN, the classification head along with the SG module is placed {\em after} the class-agnostic bounding box regression head~\cite{girshick2015fast}; thus, the whole pipeline becomes a serial structure. The classification head and the SG module use the refined boxes for feature extraction rather than the RoIs. Thus, the pooled features are strictly aligned with the refined boxes. 

A light-weight self-attention branch is added into each head, as in Fig.~\ref{fig:pipeline} (b). The output of the attention head is a discrete distribution over the $k^2$ position-sensitive grid. The position-sensitive scores are then aggregated through a weighted sum based on that distribution. There are two reasons why we introduced the self-attention in each head: 1). The self-attention helps the network to capture the semantic difference between heavily overlap-ping boxes and hence the SGE can be learned effectively. 2) we suggest that merging the position-sensitive scores by averaging (as done previous work \cite{DaiLiHeSu16}) could be sub-optimal, while adding the self-attention module helps the model to learn how to merge the score better. 
The idea of our Serial R-FCN is similar to a Cascade R-CNN~\cite{CaiVa18}. However, while Cascade R-CNN stacks multiple classification and regression heads, we here only use one regression head and one classification head, thus do not introduce an extra parameter. Although the serial structure can be used by any two-stage detector, it suits the R-FCN best since no extra operation is added, and so the computation of the refined box is nearly cost free.

Placing the classification head after the regression head can bring us another benefit: It enables us to train the classification head using a higher IoU threshold. This yields more accurate bounding boxes. Without the serial structure, setting the IoU threshold to a very high value would result in the shortage of positive samples. However, in practice, we find that simply adopting the serial structure could easily yield a network that overfits on the training data. The reason is that as training progresses, the regression head becomes more and more powerful so that the classification head cannot receive enough hard negative examples (i.e., boxes whose IoU with the ground truth box is slightly smaller than the training threshold). The result is that the model cannot distinguish these examples and true positives when the model is tested. To alleviate the overfitting problem, we propose the simple but effective
approach to add some noise to the refined bounding box so that the
classification head continues to obtain hard false examples. Formally, during training, a box $b=(x,y,w,h)$ is transformed to $b'=(x',y',w',h')$ to train the classification head and the SG module:
\begin{align}
  &x' = \sigma_x w + x,  \ y'=\sigma_y h + y,\nonumber\\
  &w' = w\cdot \mathrm{exp}(\sigma_w),  \  h' = h\cdot \mathrm{exp}(\sigma_h),
\end{align}
where $\sigma_x,\sigma_y, \sigma_w,\sigma_h$ are noise coefficient drawn from a uniform distribution $\prod_{j=1}^{j=k}{(-\zeta_k, \zeta_k)}$ where the four dimensions correspond to $x,y,w,h$ respectively. In practice we set $\zeta_x=\zeta_y=0.05$ and $\zeta_w=\zeta_h=0.2$. 

The whole pipeline is trained end-to-end with the loss functions
\begin{align}
  L_\text{total} =&\ L_\text{rpn} + \alpha L_\text{det} + \beta L_\text{\agid},
  \label{eq:total-loss}
  \\
  L_\text{rpn}=&\ L_\text{cls-anchor} + L_\text{reg-anchor},
  \label{eq:rpn-loss}\\
  L_\text{det}=&\ L_\text{cls-rbox} + L_\text{reg-roi},
  \label{eq:det-loss}\\
  L_\text{\agid} =&\ L_\text{group-rbox} + L_\text{sep-rbox}, \label{eq:sge-loss}
\end{align}
where the $ L_\text{rpn}$ is the commonly used loss to train the Region Proposal Network (RPN)~\cite{RenHeGiSu15}, $L_\text{det}$ is object detection loss~\cite{girshick2015fast} and $L_\text{\agid}$ is the loss to train \agid \ as described in Sec.~\ref{sec:eag1}. We use two hyper-parameter $\alpha$ and $\beta$ to balance between losses (Eq.~\ref{eq:total-loss}). 
The RPN classification and regression losses are applied to the anchor boxes (Eq.~\ref{eq:rpn-loss}),
the regression loss to RoIs (Eq.~\ref{eq:det-loss}), and the classification, group, and separation losses to the refined boxes (Eq.~\ref{eq:sge-loss}).

\section{Experiments}
\begin{figure}[!t]
\centering
  \includegraphics[width=0.7\columnwidth]{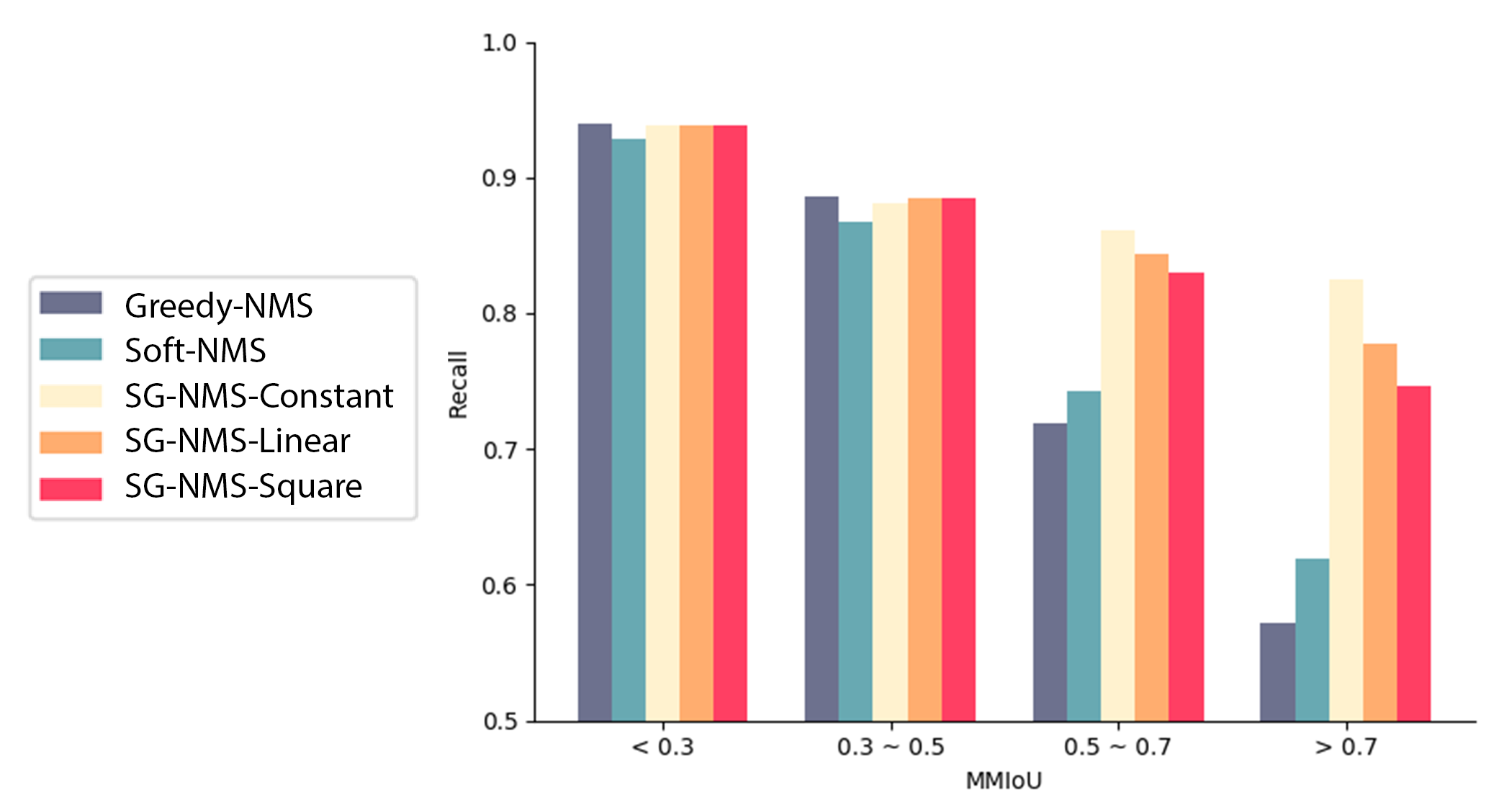}
  \caption{
  Detection recall of the proposed SG-NMS
 and competing NMS algorithms on the KITTI validation set for different levels of
    occlusion, denoted by the max-mutual-IoU (MMIOU)
    among ground-truth boxes.   }
  \label{figure5}
\end{figure}

We conducted quantitative experiments on two commonly used urban scenes datasets:  KITTI~\cite{geiger2012we} and CityPersons~\cite{zhang2017citypersons}.  To show the advantage of our {\em SG-Det model} and also to give deep insights into our approach, we first conducted several experiments using different settings on the KITTI validation set.  We then compared our approach with other state-of-the-art methods on the KITTI test set, and finally we show the performance on the CityPersons datasets. Results demonstrate the effectiveness of SG-Det to detect heavily-occluded cars and pedestrians in urban scenes. 
See also \href{http://www.cs.bu.edu/fac/betke/research/SG-Det}{www.cs.bu.edu/$\sim$betke/research/SG-Det}.

\subsection{Datasets}
\textbf{KITTI} contains 7,481 images for training and validation, and another 7,518 images for testing. We evaluated our methods on the car detection task where intra-class occlusions tend to happen the most. The dataset has a standard split into three levels of difficulty: Easy, Moderate, and Hard, according to the object scale, occlusion level, and maximum truncation. To further demonstrate how our methods handles intra-class occlusions, we proposed a new difficulty split that divide the dataset into disjoint subsets based on the max-mutual-IoU ($\mmiou$), denoted by $\mmiou$, between ground-truth boxes. The max-mutual-IoU of a ground-truth box is defined by its maximum IoU with other ground-truth boxes in the same category. We separate the validation set into three levels: Bare ($0<\mmiou \leq0.2$), Partial ( $0.2<\mmiou \leq 0.5$) and Heavy ( $0.5<\mmiou$). Average Precision (AP) is used to evaluate performance~\cite{geiger2012we}. Following prior work~\cite{ren2017accurate}, we randomly held out 3,722 images for validation and use the remaining 3,759 images for training, in which a simple image $L_2$ similarity metric was adopted to differentiate training and validation images.

\textbf{CityPersons} contains 5,000 images (2,975 for training, 500 for validation, and 1,525 for testing). The log-average Miss Rate (MR) is used to evaluate performance. Following~\cite{WangXiJiShSuSh18}, we compare the detection log-average Miss Rate (MR) in different occlusion degrees. Following prior work~\cite{WangXiJiShSuSh18}, we  separated the data into four subsets according to occlusion degree. 

\subsection{Implementation Details}
We implemented our Serial R-FCN in TensorFlow~\cite{abadi2016tensorflow} and trained it on a Nvidia Titan V GPU. For KITTI, we chose a ResNet-101~\cite{he2016deep} based on a Feature Pyramid Network (FPN) as the backbone and set the batch size to 4. The model was trained for 100,000 iterations using the Adam~\cite{kingma2014adam} optimizer with learning rate of 0.0001. For CityPersons, we chose a ResNet-50~\cite{he2016deep} as the backbone network and trained the model for 240,000 iterations with batch size of 4, and the initial learning rate was set to 0.0001 and decreased by a factor of 10 after 120,000 iterations. In all experiments, OHEM is adopted to accelerate convergence \cite{DaiLiHeSu16}. For both datasets, we set $\theta$, $\sigma$, $\rho$ to 0.7, 0.3, and 1.0, respectively, and set $\alpha$ and $\beta$ to 1.
Our code is available https://github.com/ChenhongyiYang/SG-NMS.

\subsection{Effectiveness of SG-NMS}
We report the performance of different NMS algorithms on the KITTI validation set applied to the same initial boxes so that a fair comparison is ensured
(Table~\ref{tbl-kitti-nms-official}).
For Soft-NMS, we only report the results of the linear version because we find its performance is consistently better than the Gaussian version. All three \agnms \ algorithms outperform the Greedy-NMS and Soft-NMS on the Moderate and Hard levels. In particular,  SG-NMS-Linear outperforms Greedy-NMS and Soft-NMS by $2.33$ pp and $1.39$ pp, respectively, on the Hard level where heavy intra-class occlusions occur. 
We also explored the efficacy of the Relation Network~\cite{HuGuZhDaWe18} in occlusion situations, but found that it did not work well due to generating numerous false positive detections
in crowded scenes.

\begin{table*}[t]
  \begin{center}
    \begin{tabular}{c|ccc}
      \hline
      Algorithm &  Easy & Moderate & Hard $\uparrow$ \\
      \hline
      Greedy-NMS \hspace{0.1cm} &\textbf{97.98} & 95.16 & 90.21  \\
      Soft-NMS & 97.72 & 95.13 & 91.15 \\
      \hline
      SG-Constant &  97.56 & 95.35 & 92.31 \\
      SG-Linear &  97.69 & \textbf{95.41} & \textbf{92.54} \\
      SG-Square &  97.52 & 95.14 & 92.38  \\
      \hline
    \end{tabular}
  \end{center}
  \caption{Average precision (AP) in $\%$ of the proposed SG-NMS algorithm and other commonly-used NMS algorithms on the KITTI validation set.} 
\label{tbl-kitti-nms-official}
\end{table*}

\begin{table*}[t]
  \begin{center}
  \resizebox{\textwidth}{!}{
    \begin{tabular}{cccc|cccc|cccc|cccc}
      \hline
      \multicolumn{4}{c|}{Greedy} &
      \multicolumn{4}{c|}{SG-Constant} &
      \multicolumn{4}{c|}{SG-Linear} &
      \multicolumn{4}{c}{SG-Square} \\
      $N_t$ & Bare & Partial & Heavy &  $c$ & Bare & Partial & Heavy & $t_l$ & Bare & Partial & Heavy & $t_s$ & Bare & Partial &  Heavy $\uparrow$ \\
      \hline
      soft & \textcolor{red}{\textbf{94.63}} & 84.62 & 54.62  & 1.2  & \textbf{93.77} & 84.49 & 57.21 & 2.0  & \textbf{93.74} & 85.14 & 58.72 & 3.0  & \textbf{93.65} & \textcolor{red}{\textbf{85.23}} & 60.36 \\
      0.3  & 94.33 & 76.56 & 35.10 & 1.1  & 93.73 & \textbf{84.65} & 58.74 & 1.9  & 93.72 & \textbf{85.19} & 59.17 & 2.9  & 93.62 & 85.16 & 60.24 \\
      0.4  & 94.03 & 83.38 & 40.58 & 1.0  & 93.46 & 84.33 & 60.02 & 1.8  & 93.70 & 85.11 & 59.81 & 2.8  & 93.61 & 85.14 & 60.83 \\
      0.5  & 93.63 & \textbf{85.15} & 50.63 & 0.9  & 93.58 & 84.52 & \textbf{60.05} & 1.7  & 93.68 & 85.10 & 60.19 & 2.7  & 93.59 & 85.17 & 61.48 \\
      0.6  & 91.56 & 82.85 & 55.49 & 0.8  & 93.46 & 84.33 & 60.02 & 1.6  & 93.64 & 85.06 & 60.24 & 2.6  & 93.58 & 85.09 & \textcolor{red}{\textbf{62.08}} \\
      0.7  & 46.25 & 27.24 & \textbf{57.21} & 0.7  & 93.31 & 83.83 & 59.05 & 1.5  & 93.60 & 84.97 & \textbf{61.08} & 2.5  & 93.55	& 84.98 & 62.03 \\
      \hline
    \end{tabular}}
  \end{center}
  \caption{
  AP (in $\%$) of NMS algorithms with different thresholds and occlusion levels (highest AP per level in  \textcolor{red}{red}).
  }
  \label{tbl-kitti-nms}
\end{table*}

We report the detection recall on different $\mmiou$ intervals and show the results in Fig.~\ref{figure5}. When \mmiou \ is less than $0.5$, the tested NMS algorithms achieve similar recall scores. When there is severe intra-class occlusion, i.e., $\mmiou > 0.5$, the recall of Greedy-NMS and Soft-NMS drops significantly. However, all three \agnms \ keep a relatively high recall. When $\mmiou > 0.5$, the difference in recall among the three \agnms \ algorithms is caused by the different slope of their $\Phi(\cdot)$ function. This result indicates that our \agnms \ improves the detection by promoting detection recall for objects in crowded scenes. 

We report how the hyper-parameter $t$, introduced by our \agnms, affects detection performance (Table~\ref{tbl-kitti-nms}).
Overall, the variants of \agnms\ outperform Greedy-NMS and Soft-NMS for the Heavy and Partial occlusion levels, while maintaining high performance for the Bare level. For the Heavy level, the best
result, 62.08\%, is  achieved by \ags, which is 4.87 percent points (pp) higher than the best result of Greedy-NMS and Soft-NMS.  Although Greedy-NMS can achieve an AP of $55.49\%$ for the Heavy level (when $N_t=0.6$), the AP in the Bare and Partial levels drops significantly due to the false-positive boxes it generates.

\begin{table*}[t]
  \begin{center}
    \begin{tabular}{c|ccc|ccc|ccc}
      \hline
      Model & SG & Noise & Attention & Easy & Moderate & Hard & Bare & Partial & Heavy $\uparrow$\\
      \hline
      R-FCN (FPN) & - & - & - & 95.57  & 95.08  & 88.66  & 92.41  & 81.83  & 45.96\\
      \hline
      \multirow{8}{*}{Our SG-Det}
            &   &   &   & 94.77  & 94.44  & 89.30  & 92.35  & 80.94  & 44.03  \\
            &   &   & \checkmark   & 95.84  & 94.55  & 90.10  & 93.27  & 84.87  & 47.79  \\
            &   &  \checkmark  &  & 95.04  & 95.12  & 90.01 & 93.11  & 83.28  & 43.98  \\
            & \checkmark &   &  & 94.62  & 94.50  & 89.54  & 92.45  & 81.56  & 52.62 \\
            &   & \checkmark  & \checkmark  & \textbf{97.98}  & 95.16  & 90.21  & \textbf{93.63}  & 85.15  & 50.63  \\
            & \checkmark &  \checkmark  &    &  97.80  & \textbf{95.24}  & 91.86 & 93.15  & 82.71  & 51.58  \\
            & \checkmark   &    & \checkmark  & 95.25  & 94.50  & 92.30  & 93.21  & 84.62  & 58.43  \\
            & \checkmark   & \checkmark & \checkmark  & 97.52  & 95.14  & \textbf{92.38}   & 93.59  & \textbf{85.17}  & \textbf{61.48}  \\
      \hline
    \end{tabular}
  \end{center}
  \caption{AP for different settings for the proposed SG-Det model and a baseline R-FCN model on car detection on the KITTI validation set. SG stands for \agnms; Noise stands for box noise, Attention stands for the self-attention branch used in each head.}
\label{tbl-ablation-study}
\end{table*}

\subsection{Ablation Study}

We conducted an ablation study that demonstrates how the different model components affect the overall detection performance (Table~\ref{tbl-ablation-study}). Our SG-Det model is proposed for detecting occluded objects, thus the analysis is focused on the detection of objects at the Hard difficulty (in the official split) and the Heavy occlusion level.

When the self-attention and bounding box noise are removed from our Serial R-FCN, we obtain a baseline Serial R-FCN that achieves an AP of $89.30\%$ on the Hard and $44.03\%$ on the Heavy occlusion level.  When \agnms \ is included, the detection AP on the Heavy level is improved by 8.59~pp. When the self-attention branch is added into each head, the detection AP in the Hard and Heavy levels is lifted by 0.8~pp and 3.76~pp, respectively, compared to the baseline Serial R-FCN. This verifies our assumption that the learnable score aggregation enabled by the self-attention is superior to the naive average aggregation. By adding \agnms, the APs are further improved to $92.3\%$ and $58.43\%$, which indicates that the self-attention head is important in capturing the semantic difference between heavily overlapping boxes. By adding box noise during training, the detection APs for all settings are improved, except for the heavy occlusion level. This means that the box noise can improve the detection precision by alleviating the overfitting problem in the Serial R-FCN, but it cannot help with improving the detection recall for heavily occluded objects. By combining self-attention, box noise, and \agnms ,  the full SG-Det model achieves APs of $92.38\%$ and $61.48\%$ on the Hard difficulty and Heavy occlusion level, respectively. 

To conclude, we note that self-attention is useful to capture the semantic difference between heavily overlapping boxes. The box noise can alleviate the overfitting problem so that the detection precision is improved and the \agnms \ algorithm can improve the detection performance for heavily occluded objects. 

\subsection{Discussion}

\begin{table*}[t]
  \begin{center}
    \begin{tabular}{c|ccc}
      \hline
      Embedding  & Bare & Partial & Heavy $\uparrow$ \\
      \hline
      SE & 93.31 & 83.95 & 55.60 \\
      GE & \textbf{94.11} & 78.53 & 38.12 \\  
      SGE & 93.59 & \textbf{85.14} & \textbf{61.48}  \\
      \hline
    \end{tabular}
  \end{center}
  \caption{Comparison of AP between the proposed \agidfull \ , the pure Semantic Embedding (SE) and the pure Geometric Embedding (GE) on the KITTI validation set.}
\label{tbl-ge-se}
\end{table*}

\begin{table*}[t]
  \begin{center}
    \begin{tabular}{c|c|c|c|c|c|c|c|c|c|c}
      \hline
      $\rho$ & 0.05 & 0.10 & 0.15 & 0.20 & 0.25 & 0.30 & 0.35 &0.40 & 0.45 & 0.50 \\
      \hline
      Bare  & 93.92 & 94.03 & 92.50 & \bf{94.24} & 93.28 & 93.68 & 92.98 & 93.56 & 93.17 & 93.07 \\
      Partial & 81.73 & 82.89 & 83.75 & 84.06 & \bf{85.70} & 85.10 & 84.89 & 82.38 & 83.24 & 83.52 \\
      Heavy  & 53.42 & 54.73 & 58.37 & 57.08 & 60.05 & \bf{60.19} & 58.93 & 56.05 & 55.11 & 53.25 \\
      \hline
    \end{tabular}
  \end{center}
  \caption{Comparison of AP using different $\rho$ during training.} 
\label{tbl-rho}
\end{table*}

\subsubsection{The importance of Semantics and Geometry.}
We explored the importance of the semantic and geometric features by removing them from the embedding calculation. We first removed the semantic features by computing a Geometric Embedding (GE) for each box, where the GE is computed using a fixed $\hat{s}$ that is the mean of all the $s$ vectors in the validation set. The performance of GE, shown in Table~\ref{tbl-ge-se}, is inferior than our \agid \ in occlusion situations, demonstrating the benefit of computing semantic features adaptively. Then we tested the purely-semantic model: for every box, a $1D$ Semantic Embedding (SE) is computed directly from its pooled region feature (Table~\ref{tbl-ge-se}). Our \agid \ performs better than the SE for all three occlusion levels. In fact, the two loss functions, defined in Sec.~\ref{sec:eag1} for the SE, produce very unstable results during training.  This means it is difficult for the neural network to learn such an embedding based on semantic features only, and it reveals the benefit of including geometric features.


\subsubsection{How to set $\rho$ when training \agid ?}
 We use a hyper-parameter~$\rho$ to determine occlusion during training (Sec.~\ref{sec:eag1}): For a detected box $b$, if its second largest IoU with any ground-truth box is larger than $\rho$, we assert $b$ is occluded or occludes another object. Thus, the value of $\rho$ becomes critical to the performance. In Table~\ref{tbl-rho}, we report the AP on different $\rho$ using \agl \  with $t_l=1.7$. The results show that the performance on the bare difficulty level does not depend on $\rho$, which is reasonable because our \agid \  and \agnms \  do not affect objects without occlusion. The best $\rho$ for the partial and heavy difficulty levels are $0.25$ and $0.3$, so we suggest to use a $\rho$ of $0.27$. A different value for $\rho$ leads to a decrease in performance. To explain this, we suggest that a low value of $\rho$ brings too much noise into the computing of the {\em group loss}, while a high value of $\rho$ results in the model failing to capture the semantic difference of overlapping boxes that belong to different objects. 

\begin{table*}[t]
  \begin{center}
    \begin{tabular}{c|c|rcc}
      \hline
      Model &Runtime $\downarrow$ & \hspace{0.2cm} Easy & Moderate & Hard $\uparrow$\\
      \hline
      RRC~\cite{ren2017accurate} &  \downplay{3.60} s & \downplay{95.68} & \downplay{93.40} & \downplay{87.37} \\
      SenseKITTI~\cite{yang2016craft} & \downplay{4.50} s & \downplay{94.79} & \downplay{93.17} & \downplay{84.38} \\
      SDP+RPN~\cite{yang2016exploit} &  0.40 s & 95.16 & 92.03 & 79.16 \\
      ITVD~\cite{liu2018improving} & 0.30 s & 95.85 & 91.73 & 79.31 \\
      SINet+~\cite{hu2018sinet} &  0.30 s & 94.17 & 91.67 & 78.6 \\
      Cascade MS-CNN~\cite{CaiVa18} & 0.25 s & 94.26 & 91.60 & 78.84 \\
      LTN~\cite{wang2019learning} & 0.40 s & 94.68 & 91.18 & 81.51 \\
      Aston-EAS~\cite{wei2019enhanced} & 0.24 s & 93.91 & 91.02 & 77.93 \\
      Deep3DBox~\cite{mousavian20173d} &  1.50 s & 94.71 & 90.19 & 76.82 \\
      R-FCN(FPN)~\cite{DaiLiHeSu16,lin2017feature} & 0.20 s & 93.53 & 89.35 & 79.35 \\
      \hline
      Ours & 0.20 s &  95.81 & 93.03 & 83.00 \\
      \hline
    \end{tabular}
  \end{center}
  \caption{Runtime and AP ($\%$) on the KITTI test set as reported on the KITTI leaderboard. All methods are ranked based on Moderate difficulty.}
\label{tbl-kitti-test}
\end{table*}

\subsection{Comparison with Prior Methods} 
We compared our model with other state-of-the-art models on the KITTI car
detection leaderboard (Table~\ref{tbl-kitti-test}).  Our Serial R-FCN and \agnms \ are ranked at the third place among the existing methods. The respective APs on the Moderate and Hard level are $1.00$ pp and $3.84$ pp higher than the fourth-place values \cite{yang2016exploit}. Although  RRC~\cite{ren2017accurate} and sensekitti~\cite{yang2016craft} are ranked higher than ours, the speed of our method is more than ten times faster than theirs.  A reason is that our main contribution focuses on the post-processing step rather than the detection pipeline.

\subsection{Experiments on CityPersons}
\begin{figure*}
  \begin{subfigure}[b]{0.32\textwidth}
    \includegraphics[width=\textwidth]{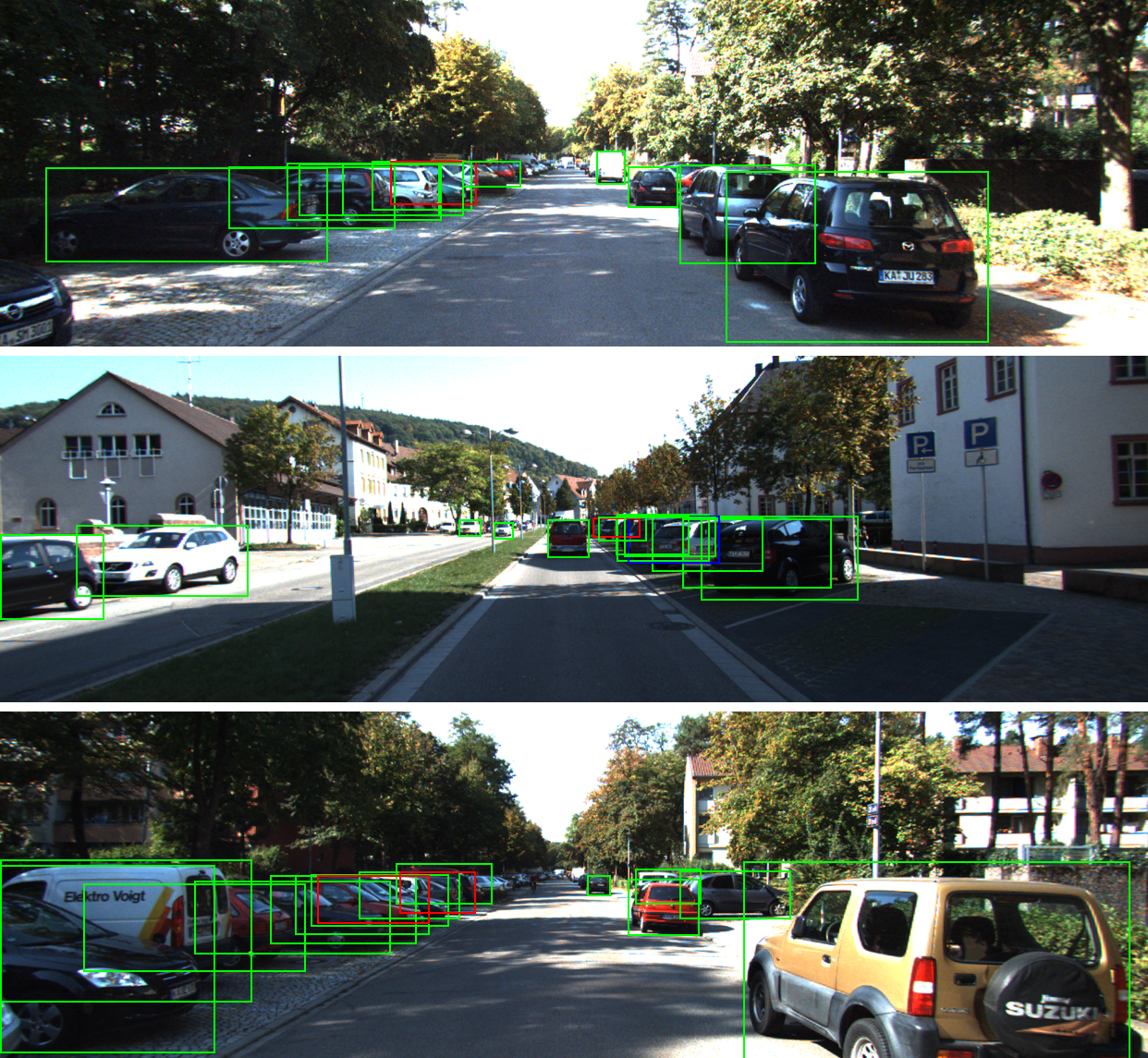}
    \caption{Proposed \agnms}
  \end{subfigure}
 \begin{subfigure}[b]{0.32\textwidth}
    \includegraphics[width=\textwidth]{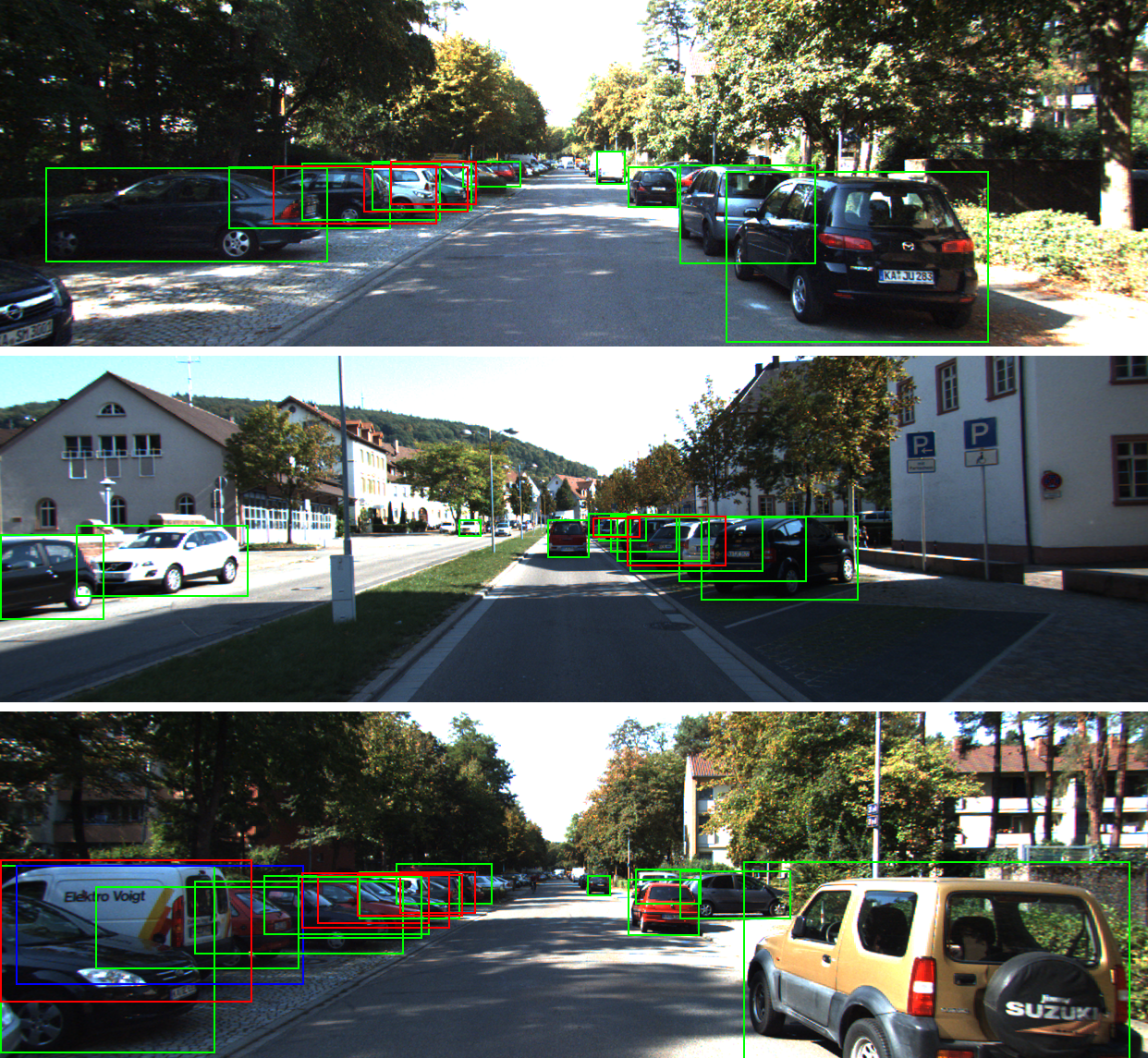}
    \caption{Greedy-NMS}
  \end{subfigure}
  \begin{subfigure}[b]{0.32\textwidth}
    \includegraphics[width=\textwidth]{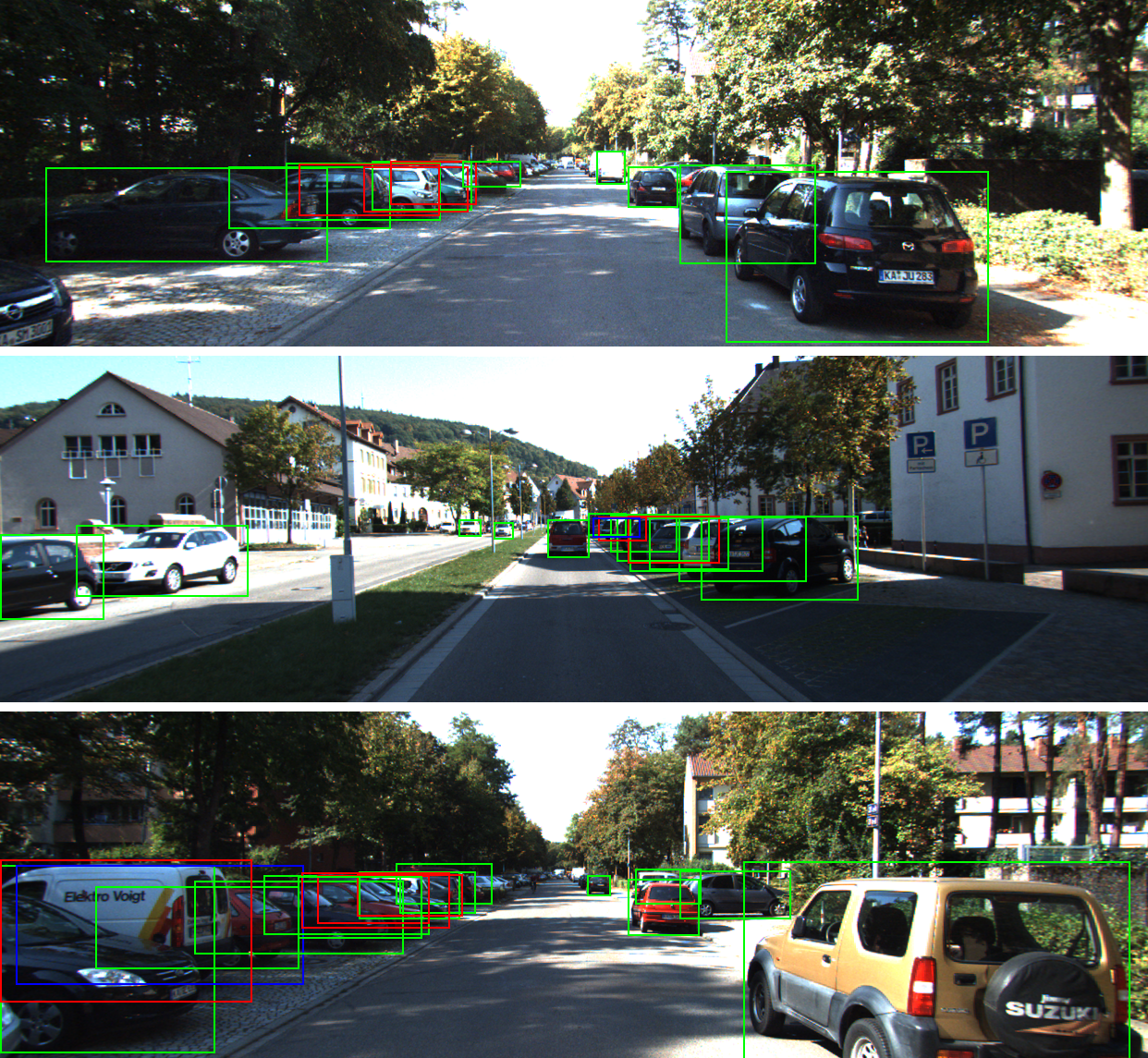}
    \caption{Soft-NMS}
  \end{subfigure}
  \caption{Visualization of results
  with true positive (\textcolor{green}{green}), false positive (\textcolor{blue}{blue}) and missed (\textcolor{red}{red}) detections.
  Two 
  failure cases of \agnms \ are shown with a false positive detection ((a) middle), and a missed detection ((a) bottom).
  }
\end{figure*}
\begin{table*}[t]
  \begin{center}
    \begin{tabular}{c|c|ccc}
      \hline
      Model & Reasonable & Bare & Partial & Heavy $\downarrow$\\
      \hline
      Adapted FasterRCNN~\cite{zhang2017citypersons} & 12.8 & - & - & - \\
      
      Repulsion Loss~\cite{WangXiJiShSuSh18} & 11.6 & 7.0 & 14.8 & 55.3 \\
      
      OR-CNN~\cite{ZhangWeBiLeLi18} & 11.0 & \textbf{5.9} &  13.7 &  51.3 \\
      
      Adaptive-NMS ~\cite{HosangBeSc17} & \textbf{10.8} &6.2 & 11.4 & 54.0 \\ 
      \hline
       SR-FCN+Greedy-NMS & 11.7 & 7.5 & 11.0 & 52.4 \\
       SerialR-FCN+Soft-NMS & 11.4 & 7.1 & 10.9 & 51.8 \\  
      \hline
      SR-FCN+SG-Constant & 11.5 & 7.4 & 11.2 & 52.3 \\
      SR-FCN+SG-Linear & 11.3 & 7.3 & 10.8 & 51.6 \\
      SR-FCN+SG-Square & 11.0 & 7.2 & \textbf{10.7} & \textbf{51.1} \\
      \hline
    \end{tabular}
  \end{center}
  \caption{The miss rate (\%) on the CityPersons validation set.}
 \label{tbl-citypersons-validation}
\end{table*}

We compare miss rates of NMS algorithms on the CityPersons validation set for different occlusion degrees in Table~\ref{tbl-citypersons-validation}. We also compare our model with existing methods. The NMS hyper-parameters are obtained from a grid search, and we report the best result for each algorithm. With Greedy-NMS, our Serial R-FCN achieves miss rates of $11.7\%$ (reasonable difficulty level) and $52.4\%$ (heavy). Using Soft-NMS yields a slight improvement. \agl \ and \ags \ yield $0.2$ and $0.7$ pp improvements (reasonable difficulty), but using \agc \ harms the performance for this level because a single threshold cannot handle the various complex occlusion situations. All three \agnms \ improve  performance on the heavy and partial occlusion levels. Especially, the \ags \ improves the respective miss rate to $10.7\%$ and $51.1\%$ on the partial and heavy occlusion levels, making our methods superior to the state of the art on those two levels. This means our method excels at handling occlusions.

\section{Conclusion}
In this paper, we presented two contributions, a novel \agidshort \ mechanism that operates on detected bounding boxes and an effective \agnmsshort\ algorithm that improves detection recall for heavily-occluded objects. Our combined model SG-Det achieves state-of-the-art performance on KITTI and CityPersons datasets by dramatically improving the detection recall and excelling in a low run time. 

\noindent
\subsubsection*{ Acknowledgements}
\noindent
We acknowledge partial support of this work by the MURI Program, N00014-19-1-2571 associated with AUSMURIB000001, and the National Science Foundation under Grant No. 1928477.

\clearpage

{\small
\bibliographystyle{ieeetr}
\bibliography{bib}

\begin{thebibliography}{10}

\bibitem{deng2009imagenet}
J.~Deng, W.~Dong, R.~Socher, L.-J. Li, K.~Li, and L.~Fei-Fei, ``Imagenet: A
  large-scale hierarchical image database,'' in {\em 2009 IEEE Conference on
  Computer Vision and Pattern Recognition}, pp.~248--255, Ieee, 2009.

\bibitem{he2016deep}
K.~He, X.~Zhang, S.~Ren, and J.~Sun, ``Deep residual learning for image
  recognition,'' in {\em Proceedings of the IEEE Conference on Computer Vision
  and Pattern Recognition}, pp.~770--778, 2016.

\bibitem{simonyan2014deep}
K.~Simonyan and A.~Zisserman, ``Very deep convolutional networks for
  large-scale image recognition,'' 2014.

\bibitem{RedmonDiGiFa16}
J.~Redmon, S.~Divvala, R.~Girshick, and A.~Farhadi, ``You only look once:
  Unified, real-time object detection,'' in {\em Proceedings of the IEEE
  conference on computer vision and pattern recognition}, pp.~779--788, 2016.

\bibitem{LiuAnErSzReFuBe16}
W.~Liu, D.~Anguelov, D.~Erhan, C.~Szegedy, S.~Reed, C.-Y. Fu, and A.~C. Berg,
  ``{SSD}: Single shot multibox detector,'' in {\em European Conference on
  Computer Vision}, pp.~21--37, Springer, 2016.

\bibitem{GirshickDoDaMa14}
R.~Girshick, J.~Donahue, T.~Darrell, and J.~Malik, ``Rich feature hierarchies
  for accurate object detection and semantic segmentation,'' in {\em
  Proceedings of the IEEE Conference on Computer Vision and Pattern
  Recognition}, pp.~580--587, 2014.

\bibitem{girshick2015fast}
R.~Girshick, ``Fast {R-CNN},'' in {\em Proceedings of the IEEE International
  Conference on Computer Vision}, pp.~1440--1448, 2015.

\bibitem{RenHeGiSu15}
S.~Ren, K.~He, R.~Girshick, and J.~Sun, ``Faster {R-CNN}: Towards real-time
  object detection with region proposal networks,'' in {\em Advances in Neural
  Information Processing Systems}, pp.~91--99, 2015.

\bibitem{BodlaSiChDa17}
N.~Bodla, B.~Singh, R.~Chellappa, and L.~S. Davis, ``Soft-{NMS}--improving
  object detection with one line of code,'' in {\em Proceedings of the IEEE
  International Conference on Computer Vision}, pp.~5561--5569, 2017.

\bibitem{newell2017associative}
A.~Newell, Z.~Huang, and J.~Deng, ``Associative embedding: End-to-end learning
  for joint detection and grouping,'' in {\em Advances in Neural Information
  Processing Systems}, pp.~2277--2287, 2017.

\bibitem{geiger2012we}
A.~Geiger, P.~Lenz, and R.~Urtasun, ``Are we ready for autonomous driving?
  {T}he {KITTI} vision benchmark suite,'' in {\em 2012 IEEE Conference on
  Computer Vision and Pattern Recognition}, pp.~3354--3361, IEEE, 2012.

\bibitem{zhang2017citypersons}
S.~Zhang, R.~Benenson, and B.~Schiele, ``Citypersons: A diverse dataset for
  pedestrian detection,'' in {\em Proceedings of the IEEE Conference on
  Computer Vision and Pattern Recognition}, pp.~3213--3221, 2017.

\bibitem{lin2017focal}
T.-Y. Lin, P.~Goyal, R.~Girshick, K.~He, and P.~Doll{\'a}r, ``Focal loss for
  dense object detection,'' in {\em Proceedings of the IEEE international
  conference on computer vision}, pp.~2980--2988, 2017.

\bibitem{DaiLiHeSu16}
J.~Dai, Y.~Li, K.~He, and J.~Sun, ``R-{FCN}: Object detection via region-based
  fully convolutional networks,'' in {\em Advances in Neural Information
  Processing Systems}, pp.~379--387, 2016.

\bibitem{CaiVa18}
Z.~Cai and N.~Vasconcelos, ``Cascade {R-CNN}: Delving into high quality object
  detection,'' in {\em Proceedings of the IEEE Conference on Computer Vision
  and Pattern Recognition}, pp.~6154--6162, 2018.

\bibitem{uijlings2013selective}
J.~R. Uijlings, K.~E. Van De~Sande, T.~Gevers, and A.~W. Smeulders, ``Selective
  search for object recognition,'' {\em International Journal of Computer
  Vision}, vol.~104, no.~2, pp.~154--171, 2013.

\bibitem{ZhangWeBiLeLi18}
S.~Zhang, L.~Wen, X.~Bian, Z.~Lei, and S.~Z. Li, ``Occlusion-aware {R-CNN}:
  detecting pedestrians in a crowd,'' in {\em Proceedings of the European
  Conference on Computer Vision (ECCV)}, pp.~637--653, 2018.

\bibitem{zitnick2014edge}
C.~L. Zitnick and P.~Doll{\'a}r, ``Edge boxes: Locating object proposals from
  edges,'' in {\em European Conference on Computer Vision}, pp.~391--405,
  Springer, 2014.

\bibitem{he2015spatial}
K.~He, X.~Zhang, S.~Ren, and J.~Sun, ``Spatial pyramid pooling in deep
  convolutional networks for visual recognition,'' {\em IEEE Transactions on
  Pattern Analysis and Machine Intelligence}, vol.~37, no.~9, pp.~1904--1916,
  2015.

\bibitem{HosangBeSc17}
J.~Hosang, R.~Benenson, and B.~Schiele, ``Learning non-maximum suppression,''
  in {\em Proceedings of the IEEE Conference on Computer Vision and Pattern
  Recognition}, pp.~4507--4515, 2017.

\bibitem{LiuHuWa19}
S.~Liu, D.~Huang, and Y.~Wang, ``Adaptive {NMS}: Refining pedestrian detection
  in a crowd,'' in {\em Proceedings of the IEEE Conference on Computer Vision
  and Pattern Recognition}, pp.~6459--6468, 2019.

\bibitem{HuGuZhDaWe18}
H.~Hu, J.~Gu, Z.~Zhang, J.~Dai, and Y.~Wei, ``Relation networks for object
  detection,'' in {\em Proceedings of the IEEE Conference on Computer Vision
  and Pattern Recognition}, pp.~3588--3597, 2018.

\bibitem{Tychsen_Smith_2018}
L.~Tychsen-Smith and L.~Petersson, ``Improving object localization with fitness
  nms and bounded {I}o{U} loss,'' {\em 2018 IEEE/CVF Conference on Computer
  Vision and Pattern Recognition}, Jun 2018.

\bibitem{HeZhWaSaZh19}
Y.~He, C.~Zhu, J.~Wang, M.~Savvides, and X.~Zhang, ``Bounding box regression
  with uncertainty for accurate object detection,'' in {\em Proceedings of the
  IEEE Conference on Computer Vision and Pattern Recognition}, pp.~2888--2897,
  2019.

\bibitem{jiang2018acquisition}
B.~Jiang, R.~Luo, J.~Mao, T.~Xiao, and Y.~Jiang, ``Acquisition of localization
  confidence for accurate object detection,'' in {\em Proceedings of the
  European Conference on Computer Vision (ECCV)}, pp.~784--799, 2018.

\bibitem{Tan_2019_ICCV}
Z.~Tan, X.~Nie, Q.~Qian, N.~Li, and H.~Li, ``Learning to rank proposals for
  object detection,'' in {\em The IEEE International Conference on Computer
  Vision (ICCV)}, October 2019.

\bibitem{WangXiJiShSuSh18}
X.~Wang, T.~Xiao, Y.~Jiang, S.~Shao, J.~Sun, and C.~Shen, ``Repulsion loss:
  Detecting pedestrians in a crowd,'' in {\em Proceedings of the IEEE
  Conference on Computer Vision and Pattern Recognition}, pp.~7774--7783, 2018.

\bibitem{zhang2018occluded}
S.~Zhang, J.~Yang, and B.~Schiele, ``Occluded pedestrian detection through
  guided attention in {CNN}s,'' in {\em Proceedings of the IEEE Conference on
  Computer Vision and Pattern Recognition}, pp.~6995--7003, 2018.

\bibitem{NohLeKiKi18}
J.~Noh, S.~Lee, B.~Kim, and G.~Kim, ``Improving occlusion and hard negative
  handling for single-stage pedestrian detectors,'' in {\em Proceedings of the
  IEEE Conference on Computer Vision and Pattern Recognition}, pp.~966--974,
  2018.

\bibitem{tian2015deep}
Y.~Tian, P.~Luo, X.~Wang, and X.~Tang, ``Deep learning strong parts for
  pedestrian detection,'' in {\em Proceedings of the IEEE international
  conference on computer vision}, pp.~1904--1912, 2015.

\bibitem{zhou2017multi}
C.~Zhou and J.~Yuan, ``Multi-label learning of part detectors for heavily
  occluded pedestrian detection,'' in {\em Proceedings of the IEEE
  International Conference on Computer Vision}, pp.~3486--3495, 2017.

\bibitem{Goldman_2019_CVPR}
E.~Goldman, R.~Herzig, A.~Eisenschtat, J.~Goldberger, and T.~Hassner, ``Precise
  detection in densely packed scenes,'' in {\em The IEEE Conference on Computer
  Vision and Pattern Recognition (CVPR)}, June 2019.

\bibitem{ZhouYu18}
C.~Zhou and J.~Yuan, ``Bi-box regression for pedestrian detection and occlusion
  estimation,'' in {\em Proceedings of the European Conference on Computer
  Vision (ECCV)}, pp.~135--151, 2018.

\bibitem{pang2019mask}
Y.~Pang, J.~Xie, M.~H. Khan, R.~M. Anwer, F.~S. Khan, and L.~Shao,
  ``Mask-guided attention network for occluded pedestrian detection,'' in {\em
  Proceedings of the IEEE International Conference on Computer Vision},
  pp.~4967--4975, 2019.

\bibitem{zhang2019double}
K.~Zhang, F.~Xiong, P.~Sun, L.~Hu, B.~Li, and G.~Yu, ``Double anchor {R-CNN}
  for human detection in a crowd,'' {\em arXiv preprint arXiv:1909.09998},
  2019.

\bibitem{law2018cornernet}
H.~Law and J.~Deng, ``Cornernet: Detecting objects as paired keypoints,'' in
  {\em Proceedings of the European Conference on Computer Vision (ECCV)},
  pp.~734--750, 2018.

\bibitem{ren2017accurate}
J.~Ren, X.~Chen, J.~Liu, W.~Sun, J.~Pang, Q.~Yan, Y.-W. Tai, and L.~Xu,
  ``Accurate single stage detector using recurrent rolling convolution,'' in
  {\em Proceedings of the IEEE Conference on Computer Vision and Pattern
  Recognition}, pp.~5420--5428, 2017.

\bibitem{abadi2016tensorflow}
M.~Abadi, A.~Agarwal, P.~Barham, E.~Brevdo, Z.~Chen, C.~Citro, G.~S. Corrado,
  A.~Davis, J.~Dean, M.~Devin, {\em et~al.}, ``Tensorflow: Large-scale machine
  learning on heterogeneous distributed systems,'' {\em arXiv preprint
  arXiv:1603.04467}, 2016.

\bibitem{kingma2014adam}
D.~P. Kingma and J.~Ba, ``Adam: A method for stochastic optimization,'' {\em
  arXiv preprint arXiv:1412.6980}, 2014.

\bibitem{yang2016craft}
B.~Yang, J.~Yan, Z.~Lei, and S.~Z. Li, ``Craft objects from images,'' in {\em
  Proceedings of the IEEE Conference on Computer Vision and Pattern
  Recognition}, pp.~6043--6051, 2016.

\bibitem{yang2016exploit}
F.~Yang, W.~Choi, and Y.~Lin, ``Exploit all the layers: Fast and accurate cnn
  object detector with scale dependent pooling and cascaded rejection
  classifiers,'' in {\em Proceedings of the IEEE Conference on Computer Vision
  and Pattern Recognition}, pp.~2129--2137, 2016.

\bibitem{liu2018improving}
W.~Liu, S.~Liao, W.~Hu, X.~Liang, and Y.~Zhang, ``Improving tiny vehicle
  detection in complex scenes,'' in {\em 2018 IEEE International Conference on
  Multimedia and Expo (ICME)}, pp.~1--6, IEEE, 2018.

\bibitem{hu2018sinet}
X.~Hu, X.~Xu, Y.~Xiao, H.~Chen, S.~He, J.~Qin, and P.-A. Heng, ``Sinet: A
  scale-insensitive convolutional neural network for fast vehicle detection,''
  {\em IEEE Transactions on Intelligent Transportation Systems}, vol.~20,
  no.~3, pp.~1010--1019, 2018.

\bibitem{wang2019learning}
T.~Wang, X.~He, Y.~Cai, and G.~Xiao, ``Learning a layout transfer network for
  context aware object detection,'' {\em IEEE Transactions on Intelligent
  Transportation Systems}, 2019.

\bibitem{wei2019enhanced}
J.~Wei, J.~He, Y.~Zhou, K.~Chen, Z.~Tang, and Z.~Xiong, ``Enhanced object
  detection with deep convolutional neural networks for advanced driving
  assistance,'' {\em IEEE Transactions on Intelligent Transportation Systems},
  2019.

\bibitem{mousavian20173d}
A.~Mousavian, D.~Anguelov, J.~Flynn, and J.~Kosecka, ``3d bounding box
  estimation using deep learning and geometry,'' in {\em Proceedings of the
  IEEE Conference on Computer Vision and Pattern Recognition}, pp.~7074--7082,
  2017.

\bibitem{lin2017feature}
T.-Y. Lin, P.~Doll{\'a}r, R.~Girshick, K.~He, B.~Hariharan, and S.~Belongie,
  ``Feature pyramid networks for object detection,'' in {\em Proceedings of the
  IEEE Conference on Computer Vision and Pattern Recognition}, pp.~2117--2125,
  2017.

\end{thebibliography}
}
\end{document}